\documentclass{article}
\usepackage{spconf,amsmath,epsfig,url}

\def\x{{\mathbf x}}

\title{Explainable Artificial Intelligence: Understanding,\\ Visualizing and Interpreting Deep Learning Models}
\name{Wojciech Samek$^1$, Thomas Wiegand$^{1,2}$, Klaus-Robert M\"uller$^{2,3,4}$\thanks{This work was supported by the German Ministry for Education and Research as Berlin Big Data Center BBDC (01IS14013A). We thank Gr{\'e}gore Montavon for his valuable comments on the paper.}}
\address{$^1$Dept.\ of Video Coding \& Analytics, Fraunhofer Heinrich Hertz Institute, 10587 Berlin, Germany\\
$^2$Dept.\ of Computer Science,Technische Universit\"at Berlin, 10587 Berlin, Germany\\
$^3$Dept.\ of Brain \& Cognitive Engineering, Korea University, Seoul 136-713, South Korea\\
$^4$Max Planck Institute for Informatics, Saarbr\"ucken 66123, Germany
}
\begin{document}
%
\maketitle
\begin{abstract} \em
With the availability of large databases and recent improvements in deep learning methodology, the performance of AI systems is reaching or even exceeding the human level on an increasing number of complex tasks. Impressive examples of this development can be found in domains such as image classification, sentiment analysis, speech understanding or strategic game playing. However, because of their nested non-linear structure, these highly successful machine learning and artificial intelligence models are usually applied in a black box manner, i.e., no information is provided about what exactly makes them arrive at their predictions. Since this lack of transparency can be a major drawback, e.g., in medical applications, the development of methods for visualizing, explaining and interpreting deep learning models has recently attracted increasing attention. This paper summarizes recent developments in this field and makes a plea for more interpretability in artificial intelligence. Furthermore, it presents two approaches to explaining predictions of deep learning models, one method which computes the sensitivity of the prediction with respect to changes in the input and one approach which meaningfully decomposes the decision in terms of the input variables. These methods are evaluated on three classification tasks.
\end{abstract}
\begin{keywords}
Artificial intelligence, deep neural networks, black box models, interpretability, sensitivity analysis, layer-wise relevance propagation
\end{keywords}

\section{Introduction}
\label{sec:intro}
The field of machine learning and artificial intelligence has progressed over the last decades.
A driving force for this development were earlier improvements in support vector machines and more recent improvements in deep learning methodology \cite{lecun2012efficient}. Also the availability of large databases such as ImageNet \cite{deng2009imagenet} or Sports1M \cite{karpathy2014large}, the speed-up gains obtained with powerful GPU cards and the high flexibility of software frameworks such as Caffe \cite{jia2014caffe} or TensorFlow \cite{abadi2016tensorflow} were crucial factors to success. Today's machine learning-based AI systems excel in a number of complex tasks ranging from the detection of objects in images \cite{he2016deep} and the understanding of natural languages \cite{cho2014learning} to the processing of speech signals \cite{deng2013new}. On top of that, recent AI\footnote{The terms artificial intelligence and machine learning are used synonymously.} systems can even outplay professional human players in difficult strategic games such as Go \cite{silver2016mastering} and Texas hold'em poker \cite{MorScience17}.
These immense successes of AI systems, especially deep learning models, show the revolutionary character of this technology, which will have a large impact beyond the academic world and will also give rise to disruptive changes in industries and societies.

However, although these models reach impressive prediction accuracies, their nested non-linear structure makes them highly non-transparent, i.e., it is not clear what information in the input data makes them actually arrive at their decisions. Therefore these models are typically regarded as {\it black boxes}. 
The 37th move in the second game of the historic Go match between Lee Sedol, a top Go player, and AlphaGo, an artificial intelligence system built by DeepMind, demonstrates the non-transparency of the AI system. AlphaGo played a move which was totally unexpected and which was commented on by a Go expert in the following way:
\begin{quotation}
{\it ``It's not a human move. I've never seen a human play this move.''} (Fan Hui, 2016).
\end{quotation}
Although during the match it was unclear why the system played this move, it was the deciding move for AlphaGo to win the game.
In this case the black box character of the AlphaGo did not matter, but in many applications the impossibility of understanding and validating the decision process of an AI system is a clear drawback. For instance, in medical diagnosis it would be irresponsible to trust predictions of a black box system by default. Instead every far reaching decision should be made accessible for appropriate validation by a human expert. Also in self-driving cars, where a single incorrect prediction can be very costly, the reliance of the model on the right features must be guaranteed. The use of explainable and human interpretable AI models is a prerequisite for providing such a guarantee.
More discussion on the necessity of explainable AI can be found in Section \ref{sec:xai}.

Not surprisingly, the development of techniques for ``opening'' black box models has recently received a lot of attention in the community \cite{DBLP:journals/jmlr/BaehrensSHKHM10, DBLP:journals/corr/SimonyanVZ13, DBLP:conf/eccv/ZeilerF14, BachPLOS15, shrikumar2016not, mahendran2016visualizing, lipton2016mythos, ribeiro2016should, zintgraf2017visualizing, doshi2017towards, MonArXiv17}. 
This includes the development of methods which help to better understand what the model has learned (i.e., it's representation) \cite{Erhan2009, mahendran2015understanding, DBLP:journals/corr/NguyenYC16} as well as techniques for explaining individual predictions \cite{DBLP:conf/cidm/LandeckerTBMKB13, DBLP:journals/corr/SimonyanVZ13, DBLP:conf/eccv/ZeilerF14, BachPLOS15, MonPR17}. A tutorial on methods from these two categories can be found in \cite{MonArXiv17}.
Note that explainability is also important for support vector machines and other advanced machine learning techniques beyond neural networks \cite{LapCVPR16}.

The main goal of this paper is to foster awareness for the necessity of explainability in machine learning and artificial intelligence.  
This is done in Section \ref{sec:xai}. After that in Section \ref{sec:methods} we present two recent techniques, namely sensitivity analysis (SA) \cite{DBLP:journals/jmlr/BaehrensSHKHM10, DBLP:journals/corr/SimonyanVZ13} and layer-wise relevance propagation (LRP) \cite{BachPLOS15}, for explaining the individual predictions of an AI model in terms of input variables.
The question of how to objectively evaluate the quality of explanations is addressed in Section \ref{sec:quality}
 and results from image, text and video classification experiments are presented in Section \ref{sec:results}.
The paper concludes with an outlook on future work in Section \ref{sec:conclusion}.

\section{Why do we need explainable AI ?}
\label{sec:xai}

The ability to explain the rationale behind one's decisions to other people is an important aspect of human intelligence.
It is not only important in social interactions, e.g., a person who never reveals one's intentions and thoughts will be most probably regarded as a ``strange fellow'', but it is also crucial in educational context, where students aim to comprehend the reasoning of their teachers. Furthermore, the explanation of one's decisions is often a prerequisite for establishing a trust relationship between people, e.g., when a medical doctor explains the therapy decision to his patient.

Although these social aspects may be of less importance for technical AI systems, there are many arguments in favor of explainability in artificial intelligence.
Here are the most important ones:

\begin{itemize}
\item {\bf Verification of the system}: As mentioned before in many applications one must not trust a black box system by default. For instance, in health care the use of models which can be interpreted and verified by medical experts is an absolute necessity. The authors of \cite{caruana2015intelligible} show an example from this domain, where an AI system which was trained to predict the pneumonia risk of a person arrives at totally wrong conclusions. The application of this model in a black box manner would not reduce but rather increase the number of pneumonia-related deaths. In short, the model learns that asthmatic patients with heart problems have a much lower risk of dying of pneumonia than healthy persons. A medical doctor would immediately recognize that this can not be true as asthma and hearth problems are factors which negatively affect the prognosis for recovery. However, the AI model does not know anything about asthma or pneumonia, it just infers from data. In this example, the data were systematically biased, because in contrast to healthy persons the majority of asthma and heart patients were under strict medical supervision. Because of that supervision and the increased sensitivity of these patients, this group has a significant lower risk of dying of pneumonia. However, this correlation does not have causal character and therefore should not be taken as basis for the decision on pneumonia therapy.

\item {\bf Improvement of the system}: The first step towards improving an AI system is to understand it's weaknesses. Obviously, it's more difficult to perform such weakness analysis on black box models than on models which are interpretable. Also detecting biases in the model or the dataset (as in the pneumonia example) is easier if one understands what the model is doing and why it arrives at it's predictions. Furthermore, model interpretability can be helpful when comparing different models or architectures. For instance, the authors of \cite{LapCVPR16, ArrACL16, ArrPLOS17} observed that models may have the same classification performance, but largely differ in terms of what features they use as the basis for their decisions. 
These works demonstrate that the identification of the most ``appropriate'' model requires explainability.
One can even claim that the better we understand what our models are doing (and why they sometimes fail), the easier it becomes to improve them.

\item {\bf Learning from the system}: Because today's AI systems are trained with Millions of examples, they may observe patterns in the data which are not accessible to humans, who are only capable of learning with a limited number of examples. When using explainable AI systems we can try to extract this distilled knowledge from the AI system in order to acquire new insights. One example of such knowledge transfer from AI system to human was mentioned by Fan Hui in the quote above. The AI system identifies new strategies to play Go, which  certainly now have also been adapted by professional human players. Another domain where information extraction from the model can be crucial are the sciences. To put it simple, physicists, chemists and biologists are rather interested in identifying the hidden laws of nature than just predicting some quantity with black box models. Thus, only models which are explainable are useful in this domain (c.f., \cite{StuJNM16, schutt2017quantum}).

\item {\bf Compliance to legislation}: AI systems are affecting more and more areas of our daily life. 
With that also legal aspects, e.g., the assignment of responsibility when the systems makes a wrong decision, have recently received increased attention.
Since it may be impossible to find satisfactory answers for these legal questions when relying on black box models, future AI systems will necessarily have to become more explainable. 
Another example where regulations may become a driving force for more explainability in artificial intelligence are individual rights.
Persons immediately affected by decisions of an AI system (e.g., persons rejected for loan by the bank) may want to know why the systems has decided in this way. 
Only explainable AI systems will provide this information.
These concerns brought the European Union to adapt new regulations which implement a ``right to explanation'' whereby a user can ask for an explanation of an algorithmic decision that was made about her or him \cite{goodman2016european}.
\end{itemize}

These examples demonstrate that explainability is not only of important and topical academic interest, but it will play a pivotal role in future AI systems.

\section{Methods for Visualizing, Interpreting and Explaining Deep Learning Models}
\label{sec:methods}
\begin{figure*}[t]
\centering
\includegraphics[width=0.75\textwidth]{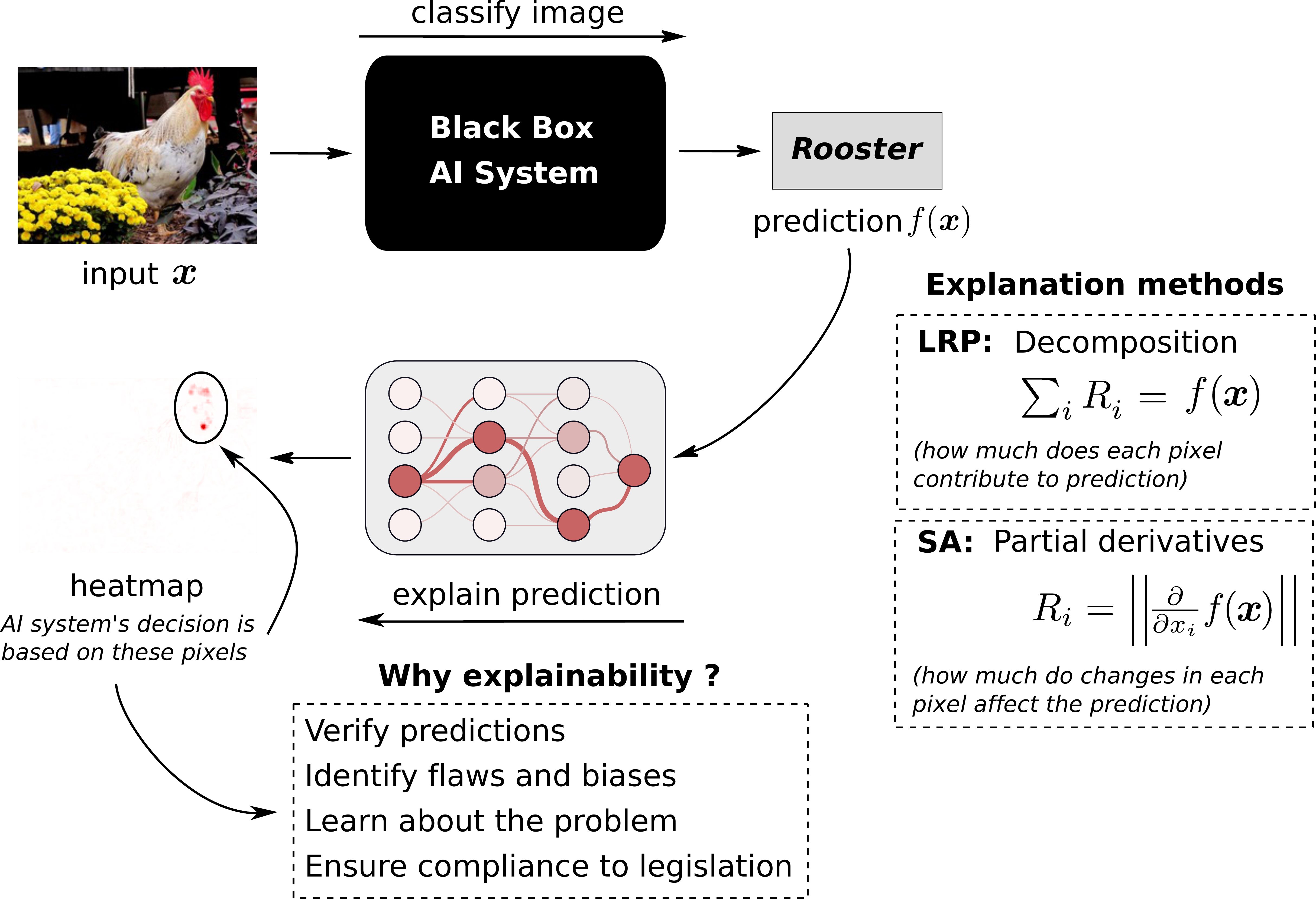}
\caption{Explaining predictions of an AI system. The input image is correctly classified as ``rooster''. In order to understand why the system has arrived at this decision, explanation methods such as SA or LRP are applied. The result of this explanation is an image, the heatmap, which visualizes the importance of each pixel for the prediction. In this example the rooster's red comb and wattle are the basis for the AI system's decision. With the heatmap one can verify that the AI system works as intended.}
\label{fig:overview}
\end{figure*}

This section introduces two popular techniques for explaining predictions of deep learning models.
The process of explanation is summarized in Fig.\ \ref{fig:overview}.
First, the system correctly classifies the input image as ``rooster''. 
Then, an explanation method is applied to explain the prediction in terms of input variables.
The result of this explanation process is a {\it heatmap} visualizing the importance of each pixel for the prediction. In this example the rooster's red comb and wattle are the basis for the AI system's decision.

\subsection{Sensitivity Analysis}
The first method is known as {\it sensitivity analysis (SA)} \cite{DBLP:journals/jmlr/BaehrensSHKHM10, DBLP:journals/corr/SimonyanVZ13} and explains a prediction based on the model's locally evaluated gradient (partial derivative). Mathematically, sensitivity analysis quantifies the importance of each input variable $i$ (e.g., image pixel) as
$$
R_i = \Big|\Big|\frac{\partial }{\partial x_i}f(\x)\Big|\Big|.
$$
This measure assumes that the most relevant input features are those to which the output is most sensitive. 
In contrast to the approach presented in the next subsection, sensitivity analysis does not explain the function value $f(\x)$ itself, but rather a {\em variation} of it.
The following example illustrates why measuring the sensitivity of the function may be suboptimal for explaining predictions of AI systems.

A heatmap computed with sensitivity analysis indicates which pixels need to changed to make the image look (from the AI system's perspective) more / less like the predicted class. For instance, in the example shown in Fig.\ \ref{fig:overview} these pixels would be the yellow flowers which occlude part of the rooster. Changing these pixels in a specific way would reconstruct the occluded parts of the rooster, which most probably would also increase the classification score, because more of the rooster would be visible in the image.
Note that such heatmap would not indicate which pixels are actually pivotal for the prediction ``rooster''.
The presence of yellow flowers is certainly not indicative of the presence of a rooster in the image.
Because of this property SA does not perform well in the quantitative evaluation experiments presented in Section \ref{sec:results}.
More discussion on the drawbacks of sensitivity analysis can be found in \cite{MonArXiv17}.

\subsection{Layer-Wise Relevance Propagation}
In the following we provide a general framework for decomposing predictions of modern AI systems, e.g., feed-forwards neural networks and bag-of-words models \cite{BachPLOS15}, long-short term memory (LSTM) networks \cite{ArrWASSA17} and Fisher Vector classifiers \cite{LapCVPR16}, in terms of input variables. In contrast to sensitivity analysis, this method explains predictions relative to the state of maximum uncertainty, i.e., it identifies pixels which are pivotal for the prediction ``rooster''. Recent work \cite{MonPR17} also shows close relations to Taylor decomposition, which is a general function analysis tool in mathematics.

A recent technique called {\it Layer-wise relevance propagation (LRP)} \cite{BachPLOS15} explains the classifier's decisions by decomposition.
Mathematically, it redistributes the prediction $f(\x)$ backwards using local redistribution rules until it assigns a relevance score $R_i$ to each input variable (e.g., image pixel).
The key property of this redistribution process is referred to as {\it relevance conservation} and can be summarized as
\begin{align}
  \sum_i R_{i} = \ldots = \sum_j R_{j} = \sum_k R_{k} = \ldots = f(\x)
    \label{eq:conservation}
\end{align}
This property says that at every step of the redistribution process (e.g., at every layer of a deep neural network), the total amount of relevance (i.e., the prediction $f(\x)$) is conserved.
No relevance is artificially added or removed during redistribution.
The relevance scores $R_i$ of each input variable determines how much this variable has contributed to the prediction.
Thus, in contrast to sensitivity analysis, LRP truly decomposes the function value $f(\x)$.

In the following we describe the LRP redistribution process for feed-forward neural networks, 
redistribution procedures have also been proposed for other popular models \cite{BachPLOS15,ArrWASSA17,LapCVPR16}.

Let $x_j$ be the neuron activations at layer $l$, $R_k$ be the relevance scores associated to the neurons at layer $l+1$ and $w_{jk}$ be the weight connecting neuron $j$ to neuron $k$. 
The simple LRP rule redistributes relevance from layer $l+1$ to layer $l$ in the following way:
\begin{align}
  R_{j} = \sum_{k}
	 \frac{x_j w_{jk}}{\sum_j x_j w_{jk} + \epsilon}R_{k}
    \label{eq:firstrule}
\end{align}
where a small stabilization term $\epsilon$ is added to prevent division by zero.
Intuitively, this rule redistributes relevance proportionally from layer $l+1$ to each neuron in layer $l$ based on two criteria, namely (i) the neuron activation $x_j$, i.e., more activated neurons receive a larger share of relevance, and (ii) the strength of the connection $w_{jk}$, i.e., more relevance flows through more prominent connections.
Note that relevance conservation holds for $\epsilon = 0$.

The ``alpha-beta'' rule is an alternative redistributes rule introduced in \cite{BachPLOS15}:
\begin{align}
  R_{j} = \sum_{k} \Big(
   \alpha\cdot \frac{(x_j w_{jk})^+}{\sum_j (x_j w_{jk})^+} -
   \beta \cdot \frac{(x_j w_{jk})^-}{\sum_j (x_j w_{jk})^-}   
  \Big) R_{k}
    \label{eq:secondrule}
\end{align}
where $()^+$ and $()^-$ denote the positive and negative parts, respectively. The conservation of relevance is enforced by an additional constraint $\alpha-\beta=1$. 
For the special case $\alpha=1$, the authors of \cite{MonPR17} showed that this redistribution rule coincides with a ``deep Taylor decomposition''  of the neural network function when the neural network is composed of ReLU neurons.

\subsection{Software}
The LRP toolbox \cite{LapJMLR16} provides a python and matlab implementation of the method as well as an integration into popular frameworks such as Caffe and TensorFlow.
With this toolbox one can directly applied LRP to other peoples' models.
The toolbox code, online demonstrators and further information can be found on \url{www.explain-ai.org}.

\section{Evaluating the Quality\\ of Explanations}
\label{sec:quality}
In order to compare heatmaps produced by different explanation methods, e.g., SA and LRP, one needs an objective measure of the quality of explanations.
The authors of \cite{SamTNNLS16} proposed such a quality measure based on perturbation analysis.
The method is based on the following three ideas:
\begin{itemize}
\item The perturbation of input variables which are highly important for the prediction leads to a steeper decline of the prediction score than the perturbation of input dimensions which are of lesser importance.
\item Explanation methods such as SA and LRP provide a score for every input variable. Thus, the input variables can be sorted according to this relevance score.
\item One can iteratively perturb input variables (starting from the most relevant ones) and track the prediction score after every perturbation step. The average decline of the prediction score (or the decline of the prediction accuracy) can be used as an objective measure of explanation quality, because
a large decline indicates that the explanation method was successful in identifying the truly relevant input variables.
\end{itemize}
In the following evaluation we use model-independent perturbations (e.g., replacing the input values by random sample from uniform distribution) in order to avoid biases.

\section{Experimental Evaluation}
\label{sec:results}
This section evaluates SA and LRP on three different problems, namely the annotation of images, the classification of text documents and the recognition of human actions in videos.

\begin{figure*}[t]
\centering
\includegraphics[width=0.85\textwidth]{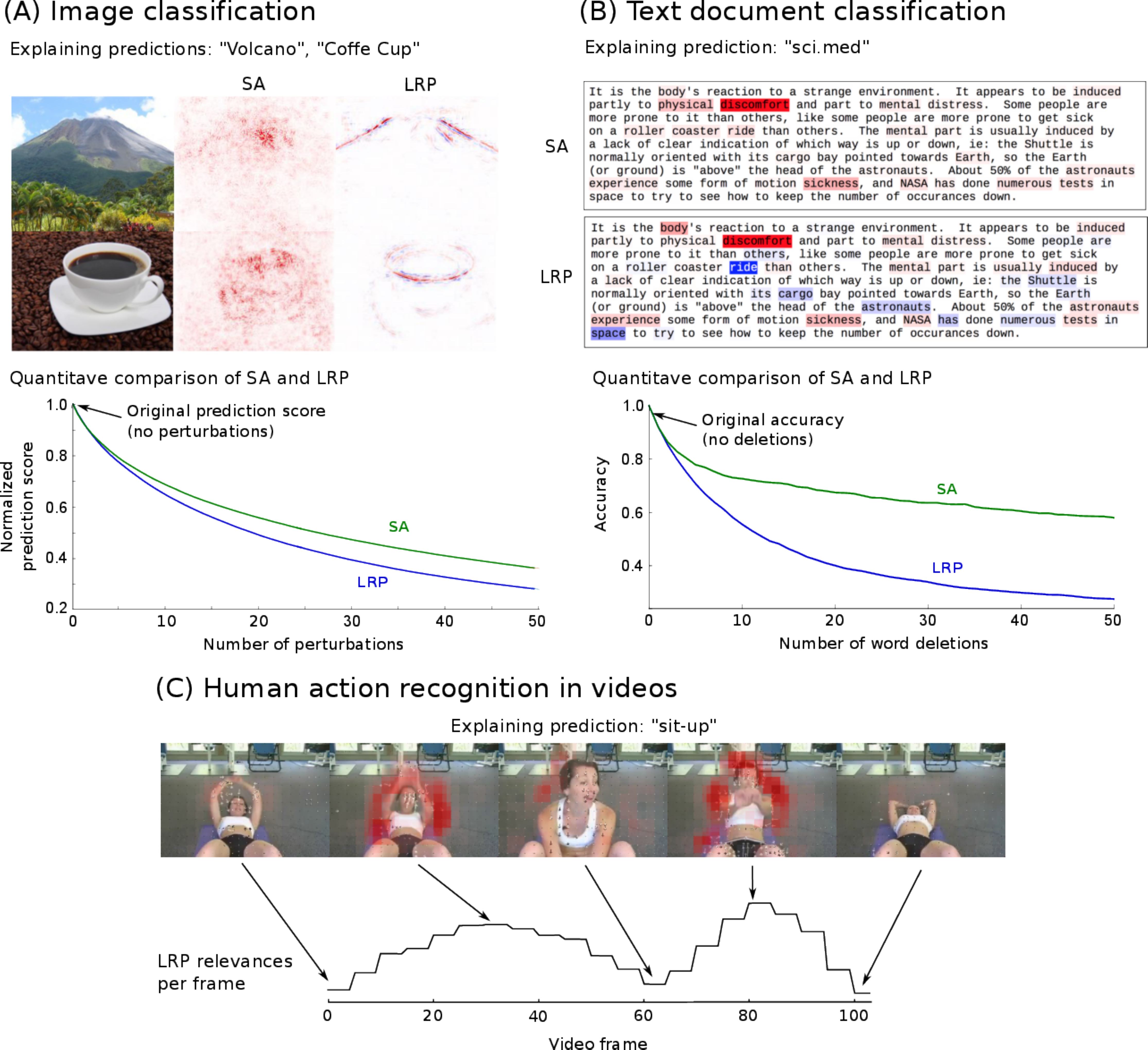}
\caption{Explaining predictions of AI systems. (A) shows the application of explainable methods to image classification. The SA heatmaps are noisy and difficult to interpret, whereas  LRP heatmaps match human intuition. 
(B) shows the application of explainable methods to text document classification. The SA and LRP heatmaps identify words such as ``discomfort'', ``body'' and ``sickness'' as the relevant ones for explaining the prediction ``sci.med''.  In contrast to sensitivity analysis, LRP distinguishes between positive (red) and negative (blue) relevances. 
(C) shows explanations for a human action recognition classifier based on motion vector features. The LRP heatmaps of a video which was classified as ``sit-up''  show increased relevance on frames in which the person is performing an upwards and downwards movement. 
}
\label{fig:results}
\end{figure*}

\subsection{Image Classification}
In the first experiment we use the GoogleNet model \cite{szegedy2015going}, a state-of-the art deep neural network, to classify general objects from the ILSVRC2012  \cite{deng2009imagenet} dataset.

Fig.\ \ref{fig:results} (A) shows two images from this dataset, which have been correctly classified as ``volcano'' and ``coffee cup'', respectively.
The heatmaps visualize the explanations obtained with SA and LRP.
The LRP heatmap of the coffee cup image shows that the model has identified the ellipsoidal shape of the cup to be a relevant feature for this image category.
In the other example, the particular shape of the mountain is regarded as evidence for the presence of a volcano in the image.
The SA heatmaps are much noisier than the ones computed with LRP and large values $R_i$ are assigned to regions consisting of pure background, e.g., the sky, although these pixels are not really indicative for image category ``volcano''. In contrast to LRP, SA does not indicate how much every pixel contributes to the prediction, but it rather measures the sensitivity of the classifier to changes in the input. Therefore, LRP produces subjectively better explanations of the model's predictions than SA.

The lower part of Fig.\ \ref{fig:results} (A) displays the results of the perturbation analysis introduced in Section \ref{sec:quality}. 
The y-axis shows the relative decrease of the prediction score average over the first  5040 images of the ILSVRC2012 dataset, i.e., a value of 0.8 means that the original scores decreased on average by 20\%. 
At every perturbation step a 9x9 patch of the image (selected according to SA or LRP scores)  is replaced by random values sampled from an uniform distribution.
Since the prediction score decrease is much faster when perturbing the images using LRP heatmaps than when using SA heatmaps, LRP also objectively provides better explanations than SA.

More discussion on this image classification experiment can be found in \cite{SamTNNLS16}.

\subsection{Text Document Classification}
In this experiment a word-embedding based convolutional neural network was trained to classify text documents from the 20Newsgroup dataset\footnote{\url{http://qwone.com/~jason/20Newsgroups}}.

Fig.\ \ref{fig:results} (B) shows SA and LRP heatmaps (e.g., a relevance score $R_i$ is assigned to every word) overlayed on top of a document, which was classified as topic ``sci.med'', i.e., the text is assumed to be about a medical topic.
Both explanation methods, SA and LRP, indicate that words such as ``sickness'', ``body'' or ``discomfort'' are the basis for this classification decision.
In contrast to sensitivity analysis, LRP distinguishes between positive (red) and negative (blue) words, i.e., words which support the classification decision ``sci.med'' and words which are in contradiction, i.e., speak for another category (e.g.,``sci.space''). Obviously, words such as ``ride'', ``astronaut'' and ``Shuttle'' strongly speak for the topic space, but not necessarily for the topic medicine.
With the LRP heatmap we can see that although the classifier decides for the correct ``sci.med'' class, there is evidence in the text which contradicts this decision.
The SA method does not distinguish between positive and negative evidence.

The lower part of the figure shows the result of the quantitative evaluation. 
The y-axis displays the relative decrease of the prediction accuracy over 4154 documents of the 20Newsgroup dataset.
At every perturbation step the most important words (according to SA or LRP score) are deleted by setting the corresponding input values to 0.
Also this result confirms quantitatively that LRP provides more informative heatmaps than SA, because these heatmaps lead to a larger decrease in classification accuracy compared to SA heatmaps.

More discussion on this text document classification experiment can be found in \cite{ArrPLOS17}.

\subsection{Human Action Recognition in Videos}
The last examples demonstrates the explanation of a Fisher Vector / SVM classifier \cite{kantorov2014efficient}, which was trained for predicting human actions from compressed videos.
In order to reduce computational costs, the classifier was trained on block-wise motion vectors (not individual pixels).
The evaluation is performed on the HMDB51 dataset \cite{kuehne2011hmdb}.

Fig.\ \ref{fig:results} (C) shows LRP heatmaps overlayed onto five exemplar frames of a video sample.
The video was correctly classified as showing the action ``sit-up''. 
One can see that the model mainly focuses on the blocks surrounding the upper body of the person.
This makes perfectly sense, as this part of the video frame shows motion which is indicative of the action ``sit-up'', namely upward and downward movements of the body.

The curve at the bottom of Fig.\ \ref{fig:results} (C) displays the distribution of relevance over (four consecutive) frames.
One can see that the relevance scores are larger for frames in which the person is performing an upwards and downwards movement.
Thus, LRP heatmaps not only visualizes the relevant locations of the action within a video frame (i.e., {\it where} relevant action happens), but it also identifies the most relevant time points within a video sequence (i.e., {\it when} relevant action happens).

More discussion on this experiment can be found in \cite{SriICASSP17}.

\section{Conclusion}
\label{sec:conclusion}
This paper approached the problem of explainability in artificial intelligence.
It was discussed why black box models are not acceptable for certain applications, e.g., in the medical domain where wrong decisions of the system can be very harmful. 
Furthermore, explainability was presented as prerequisite for solving legal questions which are arising with the increased usage of AI systems, e.g., how to assign responsibility in case of system failure. Since the ``right to explanation'' has become part of the European law, it can be expected that it will also greatly foster explainability in AI systems.

Besides being a gateway between AI and society, explainability is also a powerful tool for 
detecting flaws in the model and biases in the data, for verifying predictions, for improving models, and finally for gaining new insights into the problem at hand (e.g., in the sciences).

In future work we will investigate the theoretical foundations of explainability, in particular
the connection between post-hoc explainability, i.e., a trained model is given and the goal is to explain it's predictions,
and explainability which is incorporated directly into the structure of the model. Furthermore, we will study new ways to better understand the learned representation, especially
the relation between generalizability, compactness and explainability.
Finally, we will apply explaining methods such as LRP to new domains, e.g., communications, and search for applications of these methods beyond the ones described in this paper.


\bibliographystyle{abbrv}
\bibliography{bibliography}
\end{document}